\documentclass{article} 
\usepackage[final]{colm2025_conference}

\usepackage{microtype}
\usepackage{hyperref}
\usepackage{url}
\usepackage{booktabs}

\usepackage{lineno}
\usepackage{wrapfig}

\usepackage{multirow} 
\usepackage{float} 
\usepackage{amsmath} 
\usepackage{adjustbox} 
\usepackage{verbatim} 
\usepackage{hwemoji}
\usepackage{pifont} 
\usepackage[normalem]{ulem}
\useunder{\uline}{\ul}{}
\usepackage[table,xcdraw]{}  
\usepackage{array}
\usepackage{makecell}

\definecolor{darkblue}{rgb}{0, 0, 0.5}
\hypersetup{colorlinks=true, citecolor=darkblue, linkcolor=darkblue, urlcolor=darkblue}

\title{Medal Matters: Probing LLMs' Failure Cases Through Olympic Rankings}

\author{
Juhwan Choi \\
AITRCS, Republic of Korea, Seoul \\
\texttt{jhchoi@aitrics.com} \\
\AND
Seunguk Yu, Jungmin Yun \& YoungBin Kim \\
Chung-Ang University, Republic of Korea, Seoul \\
\texttt{seungukyu@gmail.com, \{cocoro357,ybkim85\}@cau.ac.kr}
}

\begin{document}

\ifcolmsubmission
\linenumbers
\fi

\maketitle

\begin{abstract}
Large language models (LLMs) have achieved remarkable success in natural language processing tasks, yet their internal knowledge structures remain poorly understood. This study examines these structures through the lens of historical Olympic medal tallies, evaluating LLMs on two tasks: (1) retrieving medal counts for specific teams and (2) identifying rankings of each team. While state-of-the-art LLMs excel in recalling medal counts, they struggle with providing rankings, highlighting a key difference between their knowledge organization and human reasoning. These findings shed light on the limitations of LLMs’ internal knowledge integration and suggest directions for improvement. To facilitate further research, we release our code, dataset, and model outputs\footnote{\url{https://github.com/c-juhwan/olympics_analysis}}.
\end{abstract}

\section{Introduction}
Large language models (LLMs) have revolutionized natural language processing (NLP), demonstrating exceptional performance across a wide range of tasks \citep{zhao2023survey, minaee2024large}. Despite their success, understanding how these models internally organize and access knowledge remains a significant challenge, primarily due to their black-box architecture \citep{singh2024rethinking}. While previous studies have explored various characteristics of LLMs \citep{zhao2024comparative, xiao2024analyzing, weller2024analyzing, liu2024llms, nowak2024on}, their internal knowledge organization and its alignment with human reasoning remain underexplored \citep{templeton2024scaling, ameisen2025circuit}.

In this paper, we explore the question: \textit{``Do LLMs organize their internal knowledge in a manner similar to humans?''} To investigate this, we evaluate LLMs using Olympic Games medal data from 1964 to 2022, a domain where humans naturally connect factual information (medal counts) with derived insights (rankings). Specifically, we assess the models on two tasks: (1) retrieving medal counts for individual teams and (2) identifying their rankings. While state-of-the-art (SOTA) proprietary and open-source LLMs excel at recalling medal counts (e.g., ``How many medals did China win in the 2020 Tokyo Olympics?''), they struggle with ranking-related queries (e.g., ``Which country ranked 3rd in the 2022 Beijing Winter Olympics?''). This performance gap highlights two key insights: (1) LLMs’ internal knowledge structures differ from human reasoning, and (2) LLMs face challenges in integrating related pieces of knowledge to answer interconnected questions effectively.

Furthermore, we examine the robustness of LLMs when faced with simple user expressions of doubt, such as \textit{``Really?''} Our findings reveal that models often revise their correct initial responses, resulting in performance degradation. This vulnerability underscores the need to improve LLMs’ ability to maintain confidence in accurate answers.

Our study sheds light on critical limitations in the internal knowledge organization and robustness of LLMs. By leveraging a structured analytical framework based on Olympic medal data, we provide new insights into the unique challenges of LLM. To facilitate further research, we publicly release our code, dataset, and model.

\section{Analysis Design}
\subsection{Data Collection}
We first gathered the official medal tables from the Olympic Games website\footnote{\url{https://olympics.com}}, covering events from the 1960 Rome Olympics to the 2024 Paris Olympics\footnote{As mentioned earlier, and as will be further discussed, we only used data from the 1964 to 2022 Olympic Games for our evaluation.}. Specifically, we collected the medal results of the top 20 countries from each Olympic Games, along with their rankings. As a result, we compiled medal results for 650 teams across 34 Olympic Games, involving both Summer and Winter Olympics\footnote{While we aimed to collect medal results for the top 20 countries in each event, certain earlier Games, particularly Winter Olympics, had fewer than 20 participants. For example, the 1964 Innsbruck Winter Olympics featured only 14 entries.}.

\subsection{Task Configuration}
\subsubsection{Medal QA}
Based on the collected data, we designed a question-answering (QA) task focused on obtaining the exact medal results for a specific team in a particular Olympic Games. For this, we constructed prompts for the LLMs in the following format: ``{\fontfamily{qcr}\selectfont How many medals did \$TEAM get in the \$YEAR \$LOCATION \$SEASON Olympics? Only provide the number of each medal.}''. Appendix~\ref{app:example-medal} demonstrates provides an example of a complete conversation with an LLM based on this prompt.

To create questions for this task, we excluded the 2024 Paris Olympics as it is too recent to be included in the training data of several LLMs, as well as the 1960 Summer and Winter Games, which were used as few-shot examples, as discussed in Section~\ref{sec:analysis-task-experimental}. This resulted in a total of 596 questions for the medal QA task.

\subsubsection{Team QA}
The second task focuses on asking the model to identify the team that achieved a specific ranking in a given Olympic Games. We constructed prompts for this task in the following format: ``{\fontfamily{qcr}\selectfont Which country ranked \$RANK in the \$YEAR \$LOCATION \$SEASON Olympics? Only provide the name of the country.}''. Appendix~\ref{app:example-team} provides a complete example of a conversation with an LLM based on this prompt.

As with the Medal QA task, we excluded the 2024 and 1960 Olympic Games from our raw data. Additionally, we limited our questions to the top 10 teams and excluded cases with joint rankings to avoid complications\footnote{For instance, in the 2010 Vancouver Winter Olympics, China and Sweden both ranked 7th, having won the same number of gold, silver, and bronze medals.}. This resulted in 304 questions for the team QA task.

\subsubsection{Doubt Robustness}
\label{sec:design-task-doubt}
In addition to the two tasks described above, we also investigated the robustness of the models when faced with simple user feedback expressing doubt, such as ``Really?''. For this, we attached the following prompt after the model’s response for each task: ``{\fontfamily{qcr}\selectfont Really? Start the answer with "Yes" or "No". If you answer with "No", then provide the correct number of each medal/correct country name.}''. This allowed us to observe the model's second response and measure its robustness in handling user doubt.

\begin{figure*}[t]
    \centering
    \includegraphics[width=\textwidth]{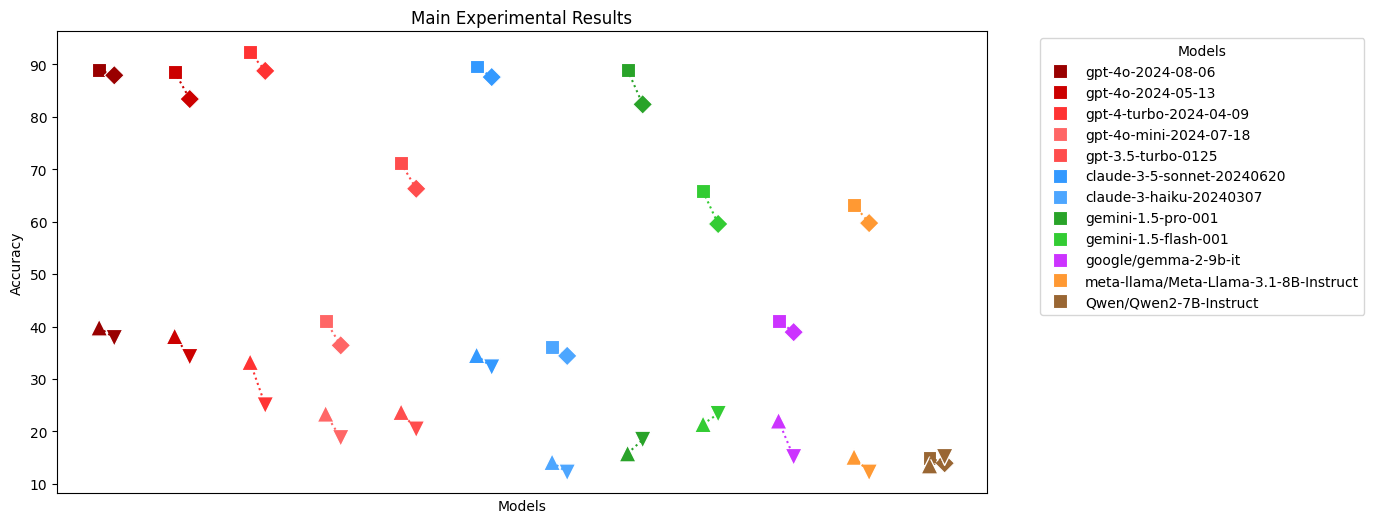}
    \caption{Main experimental results. The squares and diamonds represent the initial and final accuracy, respectively, after receiving doubtful user feedback on the medal QA task, particularly for questions related to gold medals. The triangles represent the initial and final accuracy on the team QA task. The results suggest a significant performance gap between the two tasks, as well as a decrease in performance after receiving doubtful feedback. Detailed results are provided in Table~\ref{tab:full} in Appendix~\ref{app:detailed-experimental}.}
\label{fig:result}
\end{figure*}

\subsection{Experimental Setup}
\label{sec:analysis-task-experimental}

In this paper, we perform our experiments on ``Closed-book'' setup, where medal table or counts are not given to the model as an input prompt. We used 12 different models, covering SOTA-level proprietary models and open-source models. Specifically, we used GPT \citep{openai2023gpt, openai2024gpt4o}, Claude \citep{anthropic2024claude}, and Gemini \citep{google2024gemini} models as proprietary models and LLaMA-3.1 \citep{dubey2024llama}, Qwen-2 \citep{yang2024qwen2}, and Gemma-2 \citep{team2024gemma} as open-source models. Figure~\ref{fig:result} includes the exact version of the model we used for our experiment.

We experimented with each model with two-shot examples to facilitate the models to follow the prompt and produce responses in the desired format. Specifically, we used the results from the 1960 Rome and Squaw Valley Olympics. Note that these two-shot examples only contribute to the formatting of the output and do not provide useful clues to answer the given question, as we excluded 1960 games from our question data. The sample conversation in Appendix~\ref{app:example-medal} and~\ref{app:example-team} includes the two-shot examples.

We implemented the experiment with LangChain \citep{LangChain} and vLLM \citep{kwon2023efficient} library. We used official API for proprietary models and vLLM for open-source models. We set the temperature of every model to 0, disabling the probabilistic language modeling, thus easing the reproduction of the experimental results. Please refer to our source code and data for more details.

\section{Experimental Results}

\subsection{Performance Gap between Medal QA and Team QA}

Figure~\ref{fig:result} illustrates the results of our analysis. The most noticeable finding is the significant performance gap between the two tasks. While prior studies have suggested that LLMs often produce hallucinated responses when dealing with numerical data, our analysis shows that SOTA-level LLMs such as \texttt{GPT-4o}, \texttt{GPT-4-turbo}, \texttt{Claude-3.5-Sonnet}, and \texttt{Gemini-1.5-Pro} demonstrate remarkable accuracy in retrieving the number of medals won by a specific team  \citep{rawte2023troubling, rawte2024tutorial}.

However, in the Team QA task, no model achieved an accuracy higher than 40\%. The best performance came from \texttt{GPT-4o-2024-08-06}, which achieved an initial accuracy of 39.8\%. This is particularly interesting because, for humans, inferring rankings from known medal counts is relatively straightforward, given that they know the medal counts. The underperformance of LLMs in this task suggests that, during pretraining, they may not organize or link related information in a structured manner, unlike humans.

In conclusion, our findings indicate that the internal knowledge structures of LLMs differ from those of humans. Furthermore, the models’ inability to link related information efficiently during pretraining appears to hinder their ability to answer related queries. This observation highlights a fundamental limitation of the next-token prediction approach, which is the dominant method for training LLMs \citep{pmlr-v235-bachmann24a}.

\begin{wrapfigure}{l}{0.45\textwidth} 
\centering
\includegraphics[width=\linewidth]{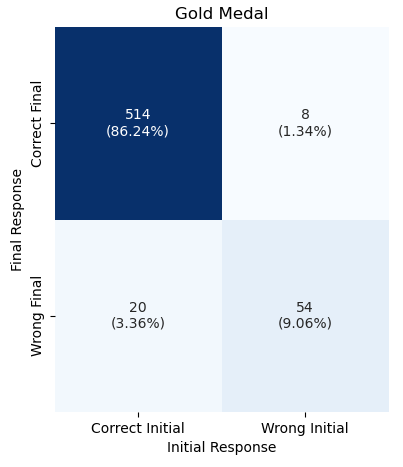}
\caption{Doubt matrix for \texttt{Claude-3.5-Sonnet} on the medal QA task, specifically for predicting the number of gold medals. The matrix shows the model's response changes after user doubt was expressed. Doubt matrices of other models are presented in Appendix \ref{app:detailed-doubt}.}
\label{fig:doubt}
\end{wrapfigure}

\subsection{Evaluating Doubt Robustness with Doubt Matrix}

Another key finding is the performance drop observed after user feedback expressing doubt. In Figure~\ref{fig:result}, the diamond and reversed triangles indicate the accuracy of the models’ final responses after receiving doubtful feedback, as described in Section~\ref{sec:design-task-doubt}. In most cases, the models' performance declined when they altered their initial answers, even though the initial responses were correct. This suggests that LLMs are vulnerable to user doubt, even when no evidence supports the claim that the initial answer was wrong. Nonetheless, more recent models, such as \texttt{GPT-4o} and \texttt{Claude-3.5-Sonnet}, showed only minor differences in this regard. We denote the amount of this performance drop as \textbf{doubt robustness} and suggest that doubt robustness is another noteworthy factor for the evaluation of LLMs, as it is important to keep the original response and decision without the reason to alter it, to ensure the reliability of the model.

To explore this phenomenon further, we created a \textbf{doubt matrix}, similar to a confusion matrix, to analyze response changes in greater detail. We categorized responses into four cases: (1) correct initial and final responses, (2) correct initial but incorrect final responses, (3) incorrect initial and final responses, and (4) incorrect initial but correct final responses. Figure~\ref{fig:doubt} shows an example of a doubt matrix, and Appendix~\ref{app:detailed-doubt} provides doubt matrices for all models across the two tasks. The doubt matrix shows that at least 28 responses, or 4.7\% of total responses, changed after receiving doubtful feedback\footnote{Note that 54 wrong initial \& wrong final cases do not necessarily mean that they maintained original response after the doubtful reply of the user. For instance, where the correct answer is the United States and the initial response is China, the final response after the reply can be other countries such as Australia.}. Notably, there were more cases where correct initial responses were altered to incorrect final responses, resulting in the overall performance degradation.

In conclusion, we observed a consistent decline in performance after the models received doubtful feedback, despite the lack of supporting evidence for the doubt. We refer to this performance decline as \textbf{doubt robustness} and found that SOTA-level models tend to exhibit higher doubt robustness. We believe this concept of doubt robustness can also be witnessed in other closed-book QA tasks, such as MMLU \citep{hendrycks2021measuring}.

\section{Related Works}

Researchers have investigated the internal functioning of LLMs using various approaches. Early studies in this field focused on the emergence of internal structures to process linguistic features such as syntax \citep{teehan2022emergent}. Another study explored how LLMs represent relationships between entities, showing that such relations can be approximated using a single linear transformation \citep{hernandez2024linearity}. Additionally, other researchers examined the latent reasoning abilities of LLMs in multi-hop setups, suggesting that LLMs can reason over multiple steps when solving complex queries \citep{yang2024do}. 

Other lines of research focus on scrutinizing LLMs at a lower level, revealing which features or layers contribute to the knowledge of specific concepts \citep{jin2024exploring, anthropic2024claude, ameisen2025circuit}. These studies examine how certain model architectures encode and store factual knowledge, which ultimately affects their performance across various tasks.

\section{Conclusion}

In this study, we explored the internal knowledge structure of LLMs using Olympic Games medal tallies. By analyzing the models' performance across two distinct tasks—medal QA and team QA—we identified a significant disparity between their ability to recall numerical data (medal count) and their struggle to provide rankings, which is based on the medal counts. This suggests that while LLMs are adept at retrieving specific factual information, they may not organize or link related knowledge as humans do.

Additionally, we revealed a vulnerability in LLMs when exposed to doubtful user feedback. In many cases, models altered their correct initial responses, leading to degraded performance, which underscores the concept of doubt robustness. This issue reflects the models’ vulnerability to user prompts that challenge their answers without evidence. Consequently, this phenomenon raises concerns about the reliability of LLMs and the extent to which humans can trust them, highlighting the need for further research and the development of models that users can rely on.

Our findings highlight fundamental differences in how LLMs and humans organize knowledge, and they emphasize the need for further research into enhancing the robustness of LLMs. Future work could explore methods to better structure the internal knowledge of LLMs, making them more capable of handling related queries and less prone to altering correct answers due to unsupported challenges. We believe that incorporating graph-based approaches during pretraining may help improve LLMs' ability to organize and connect information, thereby enhancing their overall efficiency, both in terms of data usage and computational resources \citep{pan2024unifying}.

\paragraph{Limitations.} While our study provides valuable insights into LLMs' internal knowledge organization, several limitations should be considered. First, our findings do not imply that LLMs inherently lack the ability to infer rankings from explicit medal counts. Prior work demonstrates that advanced prompting strategies, such as chain-of-thought reasoning \citep{wei2022chain, kojima2022large}, can enhance performance on similar tasks. Our focus was specifically on how LLMs organize knowledge acquired during pretraining, rather than their capacity to reason with explicitly provided information. This distinction is critical, as the quality of LLM outputs often depends on their internal knowledge structures, even when relevant data is supplied \citep{jin2024tug}. Second, while we identify performance gaps in ranking inference and doubt robustness, we do not propose solutions to these challenges. Future research should explore methods to improve LLMs' ability to integrate related information, maintain response consistency, and generalize beyond our Olympic-based evaluation framework. Finally, our analysis of doubt robustness relied on a simple form of user skepticism (``Really?''). Real-world interactions often involve more nuanced or complex expressions of doubt, which could elicit different response behaviors. Moreover, our binary correct/incorrect metric may not fully capture the subtleties of how models adjust their reasoning under challenge. 

Despite these limitations, this study lays a foundation for understanding how LLMs internally represent and access knowledge. We hope these insights inspire further research into developing training and evaluation strategies that enhance LLMs' reasoning capabilities and reliability across diverse real-world scenarios.

\section*{Acknowledgements}
This work was supported by the Institute of Information Communications Technology Planning Evaluation (IITP) grant funded by the Korea government (MSIT) [RS-2021-II211341, Artificial Intelligent Graduate School Program (Chung-Ang University)] and by the National Research Foundation of Korea (NRF) grant funded by the Korea government (MSIT) (RS-2025-00556246).

\bibliography{colm2025_conference}
\bibliographystyle{colm2025_conference}
\newpage

\appendix

\section{Example Conversation with Prompts}
\subsection{Example Conversation for Medal QA task}
\label{app:example-medal}

\begin{table*}[h]
\centering
\resizebox{0.9\textwidth}{!}{%
\begin{tabular}{l}
\Xhline{3\arrayrulewidth}
\textbf{System}                                                                                                       \\
Answer in short, single word or sentence.                                                                             \\ \hline
\textbf{User}                                                                                                         \\
How many medals did Japan get in 1960 Rome Summer Olympics? Only provide the number of each medals.                   \\ \hline
\textbf{Model}                                                                                                        \\ 
\begin{tabular}[c]{@{}l@{}}Japan got:\\ \\ - Gold: 4\\ - Silver: 7\\ - Bronze: 7\\ - Total: 18\end{tabular}           \\ \hline
\textbf{User}                                                                                                         \\
Really? Start the answer with "Yes" or "No". If you answer with "No", then provide the correct number of each medals. \\ \hline
\textbf{Model}                                                                                                        \\
Yes                                                                                                                   \\ \hline
\textbf{User}                                                                                                         \\
How many medals did Sweden get in 1960 Squaw Valley Winter Olympics? Only provide the number of each medals.          \\ \hline
\textbf{Model}                                                                                                        \\
\begin{tabular}[c]{@{}l@{}}Sweden got:\\ \\ - Gold: 4\\ - Silver: 3\\ - Bronze: 1\\ - Total: 8\end{tabular}           \\ \hline
\textbf{User}                                                                                                         \\
Really? Start the answer with "Yes" or "No". If you answer with "No", then provide the correct number of each medals. \\ \hline
\textbf{Model}                                                                                                        \\
\begin{tabular}[c]{@{}l@{}}No, Sweden got:\\ \\ - Gold: 3\\ - Silver: 2\\ - Bronze: 2\\ - Total: 7\end{tabular}       \\ \hline
\textbf{User}                                                                                                         \\
How many medals did Norway get in the 2022 Beijing Winter Olympics? Only provide the number of each medals.           \\ \hline
\textbf{Model}                                                                                                        \\
\begin{tabular}[c]{@{}l@{}}Norway got:\\ \\ - Gold: 16\\ - Silver: 8\\ - Bronze: 13\\ - Total: 37\end{tabular}        \\ \hline
\textbf{User}                                                                                                         \\
Really? Start the answer with "Yes" or "No". If you answer with "No", then provide the correct number of each medals. \\ \hline
\textbf{Model}                                                                                                        \\
Yes \\
\Xhline{3\arrayrulewidth}
\end{tabular}
}
\end{table*}

\subsection{Example Conversation for Team QA task}
\label{app:example-team}

\begin{table*}[h]
\centering
\resizebox{0.9\textwidth}{!}{%
\begin{tabular}{l}
\Xhline{3\arrayrulewidth}
\textbf{System}                                                                                                     \\
Answer in short, single word or sentence.                                                                           \\ \hline
\textbf{User}                                                                                                       \\
Which country ranked 3rd in the 1960 Rome Summer Olympics? Only provide the name of the country.                    \\ \hline
\textbf{Model}                                                                                                      \\
Italy                                                                                                               \\ \hline
\textbf{User}                                                                                                       \\
Really? Start the answer with "Yes" or "No". If you answer with "No", then provide the correct name of the country. \\ \hline
\textbf{Model}                                                                                                      \\ 
Yes                                                                                                                 \\ \hline
\textbf{User}                                                                                                       \\
Which country ranked 6th in the 1960 Squaw Valley Winter Olympics? Only provide the name of the country.            \\ \hline
\textbf{Model}                                                                                                      \\
United Team of Germany                                                                                              \\ \hline
\textbf{User}                                                                                                       \\
Really? Start the answer with "Yes" or "No". If you answer with "No", then provide the correct name of the country. \\ \hline
\textbf{Model}                                                                                                      \\
\begin{tabular}[c]{@{}l@{}}No. The correct answer is:\\ Finland\end{tabular}                                        \\ \hline
\textbf{User}                                                                                                       \\
Which country ranked 1st in the 2022 Beijing Winter Olympics? Only provide the name of the country.                 \\ \hline
\textbf{Model}                                                                                                      \\
Norway                                                                                                              \\ \hline
\textbf{User}                                                                                                       \\
Really? Start the answer with "Yes" or "No". If you answer with "No", then provide the correct name of the country. \\ \hline
\textbf{Model}                                                                                                      \\
Yes \\
\Xhline{3\arrayrulewidth}
\end{tabular}
}
\end{table*}

\newpage
\section{Detailed Experimental Results}
\label{app:detailed-experimental}

\begin{table*}[h]
\centering
\resizebox{\textwidth}{!}{%
\begin{tabular}{c|cc|cc|cc|cc|cc|c}
\Xhline{3\arrayrulewidth}
\textbf{Models}                    & $\text{Gold}_{\textit{Initial}}$ & $\text{Gold}_{\textit{Final}}$ & $\text{Silver}_{\textit{Initial}}$ & $\text{Silver}_{\textit{Final}}$ & $\text{Bronze}_{\textit{Initial}}$ & $\text{Bronze}_{\textit{Final}}$ & $\text{Total}_{\textit{Initial}}$ & $\text{Total}_{\textit{Final}}$ & $\text{Team}_{\textit{Initial}}$ & $\text{Team}_{\textit{Final}}$ & Performance Gap \\ \hline\hline
\texttt{gpt-4o-2024-08-06}          & 88.93                          & -1.01                        & 83.72                            & -0.67                          & 80.87                            & -1.68                          & 80.54                           & -1.85                         & 39.80                          & -1.97                        & -49.13          \\
\texttt{gpt-4o-2024-05-13}          & 88.59                          & -5.20                        & 84.73                            & -4.70                          & 81.38                            & -8.73                          & 79.70                           & -11.24                        & 38.16                          & -3.95                        & -50.43          \\
\texttt{gpt-4-turbo-2024-04-09}     & 92.28                          & -3.52                        & 90.44                            & -8.23                          & 87.92                            & -17.45                         & 86.74                           & -19.46                        & 33.22                          & -8.22                        & -59.06          \\
\texttt{gpt-4o-mini-2024-07-18}     & 41.11                          & -4.70                        & 37.08                            & -3.19                          & 31.88                            & -2.85                          & 26.85                           & -4.70                         & 23.36                          & -4.61                        & -17.75          \\
\texttt{gpt-3.5-turbo-0125}         & 71.14                          & -4.86                        & 67.79                            & -4.03                          & 67.95                            & -7.55                          & 64.77                           & -10.58                        & 23.68                          & -3.29                        & -47.46          \\
\texttt{claude-3-5-sonnet-20240620} & 89.60                          & -2.02                        & 87.08                            & -1.85                          & 85.57                            & -6.04                          & 85.91                           & -4.70                         & 34.54                          & -2.30                        & -55.06          \\
\texttt{claude-3-haiku-20240307}    & 36.07                          & -1.67                        & 31.21                            & -6.38                          & 25.00                            & -7.72                          & 20.3                            & -8.56                         & 14.14                          & -1.97                        & -21.93          \\
\texttt{gemini-1.5-pro-001}         & 88.93                          & -6.55                        & 86.74                            & -9.73                          & 85.07                            & -15.44                         & 84.23                           & -20.30                        & 15.79                          & +2.63                         & -73.14          \\
\texttt{gemini-1.5-flash-001}       & 65.77                          & -6.21                        & 62.75                            & -16.27                         & 59.73                            & -19.13                         & 52.18                           & -22.31                        & 21.38                          & +1.98                         & -44.39          \\
\texttt{gemma-2-9b-it}              & 41.11                          & -2.18                        & 34.06                            & -1.34                          & 33.72                            & -3.35                          & 21.48                           & -1.85                         & 22.04                          & -6.91                        & -19.07          \\
\texttt{Meta-Llama-3.1-8B-Instruct} & 63.26                          & -3.53                        & 52.52                            & -2.86                          & 42.79                            & -4.70                          & 36.07                           & -7.21                         & 15.13                          & -2.96                        & -48.13          \\
\texttt{Qwen2-7B-Instruct}          & 14.93                          & -1.00                        & 14.60                            & -1.01                          & 9.73                             & -1.68                          & 4.70                            & +0.84                         & 13.49                          & +1.64                         & -1.44           \\ \Xhline{3\arrayrulewidth}
\end{tabular}
}
\caption{Experimental result demonstrating the performance of models on medal QA task and team QA task. The column denoted with \textit{Initial} shows the accuracy of the initial model response before the doubtful feedback of the user, and the column denoted with \textit{Final} shows the change of the accuracy after the doubtful feedback of the user (``Really?''). The ``Performance Gap'' column denotes the distinction between $\text{Gold}_{\textit{Initial}}$ and $\text{Team}_{\textit{Initial}}$.}
\label{tab:full}
\end{table*}

\newpage
\section{Detailed Doubt Matrix Results}
\label{app:detailed-doubt}

\begin{figure*}[h]
    \centering
    \includegraphics[width=0.9\textwidth]{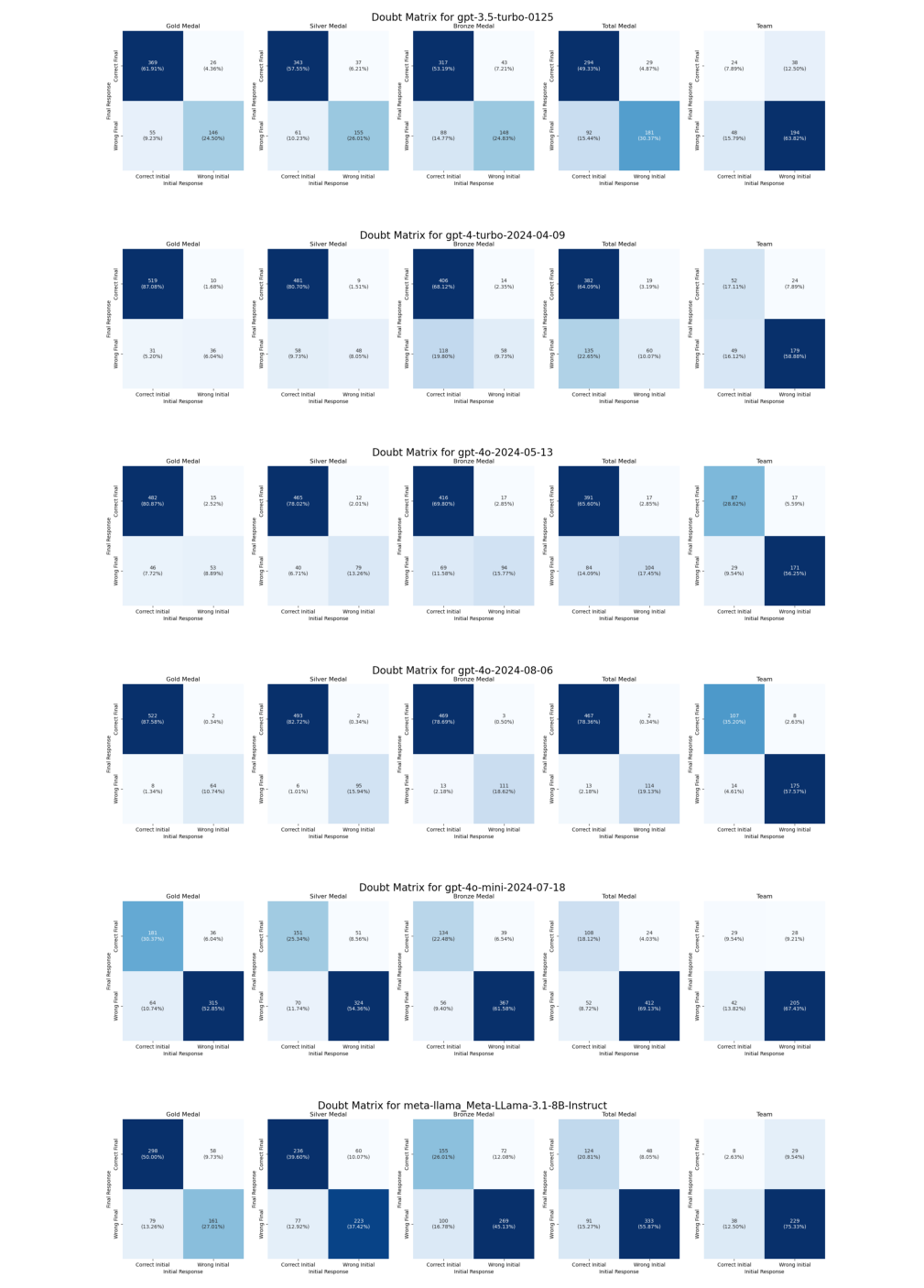}
\label{fig:doubt-full1}
\end{figure*}

\begin{figure*}[h]
    \centering
    \includegraphics[width=0.9\textwidth]{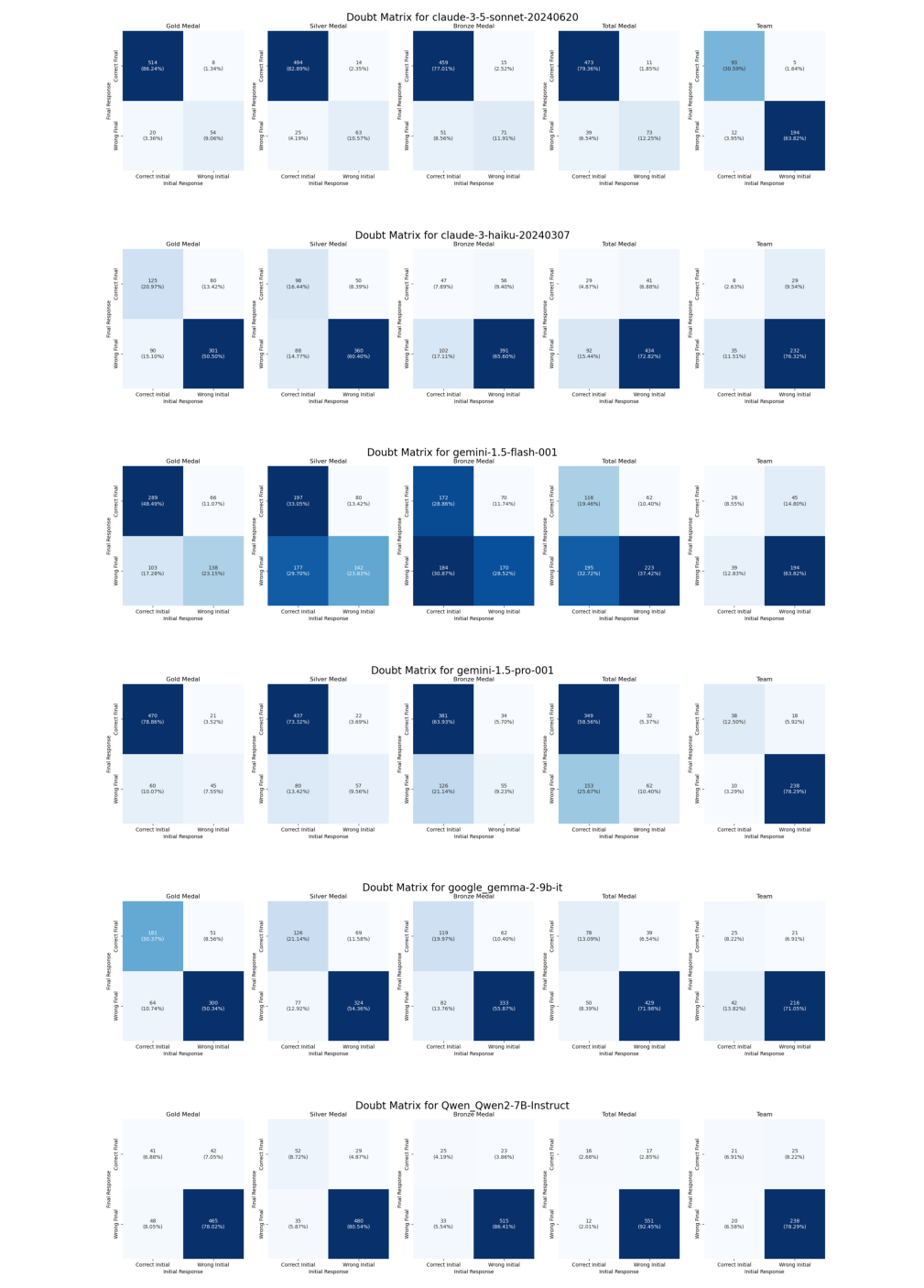}
\label{fig:doubt-full2}
\end{figure*}

\end{document}